\documentclass{article}

\usepackage[dblblindworkshop, final]{neurips_2026}
\workshoptitle{Foundation Models for Temporal Systems}

\usepackage[utf8]{inputenc} 
\usepackage[T1]{fontenc}    
\usepackage{hyperref}       
\usepackage{url}            
\usepackage{booktabs}       
\usepackage{amsfonts}       
\usepackage{nicefrac}       
\usepackage{graphicx}
\usepackage{microtype}      

\usepackage{xcolor}
\usepackage{amsmath,amssymb,amsthm}
\usepackage{tikz}
\usetikzlibrary{arrows.meta,decorations.pathreplacing,patterns}

\title{ RiverFormer: Attention Based Operational Streamflow Forecasting}

\author{%
  Atahan Ozer, Lennart Papp, Noelia Otero, and Jackie Ma \\
 Applied Machine Learning Group, Fraunhofer Heinrich-Hertz Institute,\\
 Berlin, Germany
}

\begin{document}

\maketitle

\begin{abstract}
Operational streamflow forecasting requires models to translate historical
 meteorological conditions and future weather forecasts into discharge
predictions. An important distinction from conventional time-series forecasting is that
observed discharge is unavailable as an initial condition at inference time.
Models must instead infer the evolving catchment state from historical
meteorological forcings and static attributes by learning rainfall--runoff
dynamics rather than extrapolating the observed target series. We introduce RiverFormer, an
end-to-end, fully parallel Transformer architecture for this task. RiverFormer
uses temporal patch tokenization to encode 365 days of historical
meteorological forcings and conditions on seven days of numerical weather
forecasts. It predicts the initial and complete seven-day discharge trajectory
in a single forward pass, without recurrent state updates or autoregressive
feedback between forecast leads. We train RiverFormer using a global set of 
gauges and compare it against published global
operational baselines. RiverFormer matches or outperforms the recurrent baselines
across forecast leads and geographic regions, demonstrating that fully
parallel attention is a practical alternative for large-scale operational
forecasting. By reducing temporal sequence length through patching and
eliminating sequential decoding, RiverFormer provides a scalable foundation
for future spatially distributed models operating directly on full
meteorological grids.
\end{abstract}

\section{Introduction}

Rainfall--runoff dynamics are central to understanding how catchments transform atmospheric forcing into streamflow. This process can be viewed as an energy state transition between the atmosphere and the land surface, where precipitation is partitioned into evapotranspiration, infiltration, and runoff, ultimately feeding rivers and groundwater systems. Traditional physics-based hydrological models are computationally expensive, hard to calibrate, and difficult to scale, with skill that varies substantially across basins~\citep{Horton2022}. Data-driven approaches reframe streamflow prediction as a multi-basin regression problem, with large-sample models learning transferable hydrological behaviours across many basins~\citep{kratzert2019towards,kratzert2024never}.  Current operational models mainly rely on recurrent architectures~\citep{nearing2024global, taccari2026aifl}, as their state-based design provides an inductive bias that fits the sequential and partly Markovian nature of river discharge prediction~\citep{liu2023probing}. Several studies have explored Transformer and hybrid Transformer architectures for rainfall runoff modelling~\citep{orozcolopez2024interpretable, yin2022rrformer} in different contexts. However, modern attention-based networks have not yet been fully tested in operational settings, where models must use operational weather forecasts to predict river discharge at observation sites. In this work, we provide empirical evidence that an end-to-end, fully parallel Transformer can match or outperform recurrent models for operational streamflow forecasting when trained under comparable data and compute constraints and equipped with appropriate hydrological inductive biases. We further analyze the hydrological behaviour learned by RiverFormer and contrast it with that of existing operational recurrent architectures.

\section{Related Work}

\textbf{Numerical River Discharge Forecasting.}
Rainfall-runoff modelling has long been approached as a physically grounded dynamical system, from conceptual storage-based formulations such as HBV~\citep{lindstrom1997hbv96} to spatially distributed land-surface and hydrological models such as VIC~\citep{liang1994} and mHM~\citep{samaniego2010}. Global operational systems such as GloFAS~\citep{alfieri2013glofas} couple a land-surface model with channel routing driven by ECMWF IFS forcings, producing spatially continuous discharge that encodes large-scale rainfall-runoff physics but inherits systematic biases from numerical error propagation, lack of in-situ assimilation, and unmodelled human regulation~\citep{Harrigan2020,Harrigan2023}.

\begin{figure*}[h]
    \centering
    \includegraphics[width=0.8\textwidth]{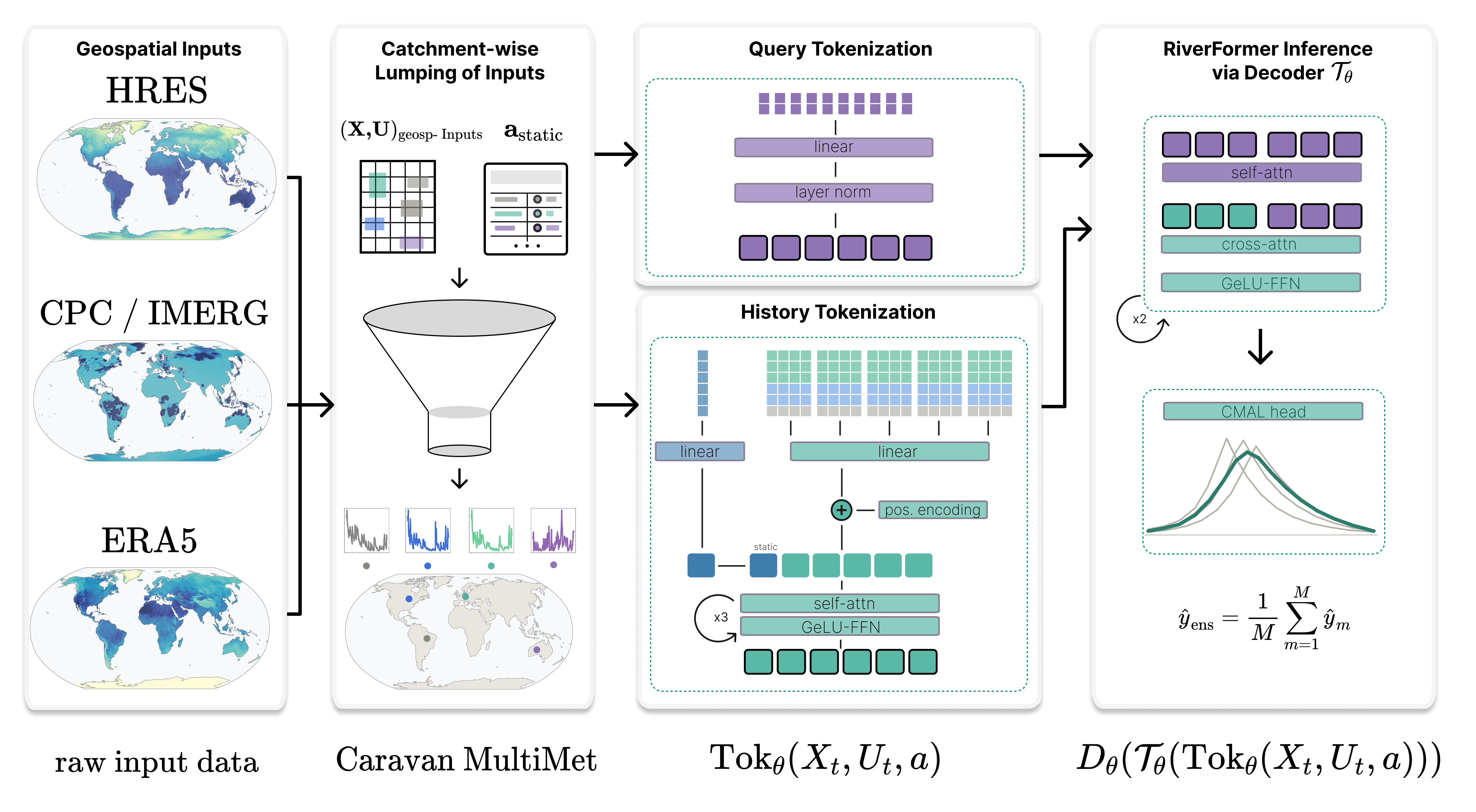}
    \caption{Overview of the RiverFormer pipeline from raw meteorological inputs to probabilistic streamflow forecasts.}
    \label{fig:main}
\end{figure*}

\textbf{ML for River Discharge.}
Station-level discharge prediction fits the sequence-to-sequence paradigm. Early applications of recurrent neural
networks (RNNs) to rainfall--runoff modelling~\citep{carriere1996virtual,hsu1997recurrent}
were followed by LSTM-based rainfall--runoff
models~\citep{hochreiter1997lstm,kratzert2018,kratzert2019towards,hoedt2021mc}, typically reducing the spatio-temporal problem through \textit{lumping}: catchment-averaged forcings and static attributes become per-station time series (Fig.~\ref{fig:main}). Operational forecasting additionally replaces future observations with weather forecasts. This setting has been demonstrated regionally~\citep{hunt2022using} and globally by the ED-LSTM of~\citet{nearing2024global} and AIFL~\citep{taccari2026aifl}. Recent non-lumped models instead operate directly on gridded forcing fields~\citep{vischer2025spatially,RiverMamba,yang2025global}, typically at greater computational cost and limited operational and regional coverage.

In contrast, sequence modeling more broadly has shifted toward parallelizable architectures such as Transformers and modern state-space models~\citep{gu2024mamba,vaswani2017attention,yang2025gated}. \citet{liu2023probing} argue that LSTMs retain a favorable Markovian inductive bias for rainfall--runoff dynamics, while Transformer-based studies~\citep{yin2022rrformer,orozcolopez2024interpretable,ambika2025novel} remain regional or non-global in their operational scope. No fully attention-based model has yet been demonstrated as a global operational neural river forecaster comparable to~\cite{nearing2024global} and \cite{taccari2026aifl}. RiverFormer aims to bridge this gap with a fully parallel temporal architecture that can naturally serve as a foundation for future grid-based extensions through spatio-temporal attention.

\section{Task, Data, and Baselines}
\label{sec:data}

\textbf{Task.}
We formulate streamflow forecasting as a temporally split, multi-horizon
regression task. At issue time $t$, the model receives the previous $K=365$
days of meteorological forcings $X_t\in\mathbb{R}^{K\times 9}$, weather
forecasts for the next seven days $U_t\in\mathbb{R}^{7\times 6}$, and static
catchment attributes $a\in\mathbb{R}^{84}$. It jointly predicts the discharge at issue time
and seven forecast lead times:
\[
F_\theta(X_t,U_t,a)
=
\left(\widehat{q}_{t+\tau}\right)_{\tau=0}^{7},
\]
where $\tau=0$ denotes the nowcast and $\tau=1,\ldots,H$ the forecast leads. 
Training and evaluation are separated in time but use the same gauges, and
only information available at issue time is provided.

\textbf{Data.}
We use $3{,}616$ gauges from Caravan and Caravan-MultiMet
\citep{kratzert2023caravan,shalev2024caravanmultimetextendingcaravan}. Following
\citet{nearing2024global,taccari2026aifl}, the historical meteorological
sequence uses ECMWF HRES lead-one forecasts; before the HRES archive begins in
2016, these inputs are replaced by ERA5-Land. Further details are provided in
Appendix~\ref{app:data}.

\textbf{Baselines and comparison protocol.}
We distinguish a controlled architecture comparison from published benchmarks.
\textbf{ED-LSTM (matched)} is our three-seed retraining of the encoder--decoder
LSTM of~\citet{nearing2024global}, using the same gauges, inputs, temporal split,
probabilistic head, training budget, and checkpoint rule as RiverFormer.
\textbf{Google v1 (published)} denotes the released forecasts of that study,
whose model was trained on more gauges, includes snowfall, and uses temporal
cross-validation. \textbf{GloFAS} is the v4.0 reanalysis of the physics-based
land-surface and routing system~\citep{alfieri2013glofas,reanalysis4_0};  it is a reference line rather than a forecast baseline.
\textbf{RiverMamba}~\citep{RiverMamba} is a flood-oriented state-space
model that upweights extremes; because it consumes that same GloFAS reanalysis
discharge, we treat it as a complementary flood benchmark rather than a fully
operational baseline. Architectural conclusions therefore rely on the matched
ED-LSTM; the other systems provide external context. Full protocols are given in
Appendix~\ref{app:training}.

\section{Method}
\label{sec:method}

\textbf{RiverFormer.}
RiverFormer maps the complete forecasting input directly to the discharge trajectory:

$$
\widehat{Y}_t =
D_\theta\!\left(
\mathcal{T}_\theta\!\left(
\operatorname{Tok}_\theta(X_t,U_t,a)
\right)\right),
$$

where $X_t$ denotes the historical forcings, $U_t$ the future weather forecasts, and $a$ the static catchment attributes defined in Section~\ref{sec:data}, respectively. $\operatorname{Tok}_\theta$ constructs the historical and forecast tokens, $\mathcal{T}_\theta$ denotes the Transformer encoder--decoder, and $D_\theta$ is the probabilistic discharge head. All $H+1$
outputs are produced in a single parallel decoding pass, without feeding previous discharge predictions back into the model.

\textbf{Patch-based tokenization.}
Unlike prior Transformer rainfall--runoff models, which predominantly treat individual hydrological timesteps as tokens~\citep{yin2022rrformer,liu2023probing,orozcolopez2024interpretable}, RiverFormer represents the $K$-day historical sequence using non-overlapping temporal patches, following modern time-series models~\citep{nie2023patch-tst,das2024timesfm,ansari2025chronos2}. Let $p$ be the patch length and $P=\lfloor K/p\rfloor$ the number of complete patches. For a historical patch $X_t^{(i)}\in\mathbb{R}^{p\times 9}$, the corresponding token is

$$
h_i =
W_p\operatorname{vec}\!\left(X_t^{(i)}\right)
+e_i,
\qquad i=1,\ldots,P ,
$$

where $\operatorname{vec}(X_t^{(i)})\in\mathbb{R}^{9p}$ flattens the $i$-th patch, $W_p$ defines the learned patch projection, and $e_i$ is a
learned positional embedding, respectively. With $K=365$ and $p=4$, RiverFormer operates on $P=91$ complete temporal patches rather than $365$ daily tokens. Static attributes $a$ are compressed by an MLP and prepended as one additional context token. Patching therefore reduces the attention sequence length while providing a local temporal inductive bias.

\textbf{Encoder and decoder.}
The static and historical tokens are processed by a Transformer encoder, producing a sequence of contextualized representations that we denote by $M_t$. Unlike a recurrent encoder that represents the complete lookback through a single hidden state, RiverFormer retains the individual encoded patch states, allowing the decoder to attend selectively to different periods of the history. Future weather forecasts enter through one decoder query per forecast lead. Let $u_{t,\tau}$ denote the weather forecast available at issue time $t$ and valid at lead $\tau$, including its availability indicator. For $\tau=1,\ldots,H$, the corresponding decoder query is $
z_\tau =
W_u u_{t,\tau}
+\ell_\tau ,
$ where $W_u$ projects the forecast variables into the decoder representation space and $\ell_\tau$ is a learned embedding identifying the forecast lead time. For $\tau=0$, which corresponds to a nowcast, a trainable nowcast embedding is used because it has no corresponding future-weather input. The $H+1$ queries are decoded jointly against $M_t$. Self-attention among the queries models dependencies across forecast horizons, while decoder cross-attention allows each lead to retrieve relevant conditions. 
All leads are predicted simultaneously without recursion, avoiding error propagation across the forecast horizon. Each decoded state parametrizes a three-component countable mixture of asymmetric Laplace distributions (CMAL)~\citep{klotz2022uncertainty,nearing2024global}. Training minimizes the negative log-likelihood of the observed discharge
under these CMAL distributions, averaged over the seven forecast
leads; details of the probabilistic head and architectural configuration are given in Appendix Table~\ref{tab:riverformer-config}. 

\section{Results and Discussion}
\label{sec:results}

\textbf{Controlled comparison with ED-LSTM.}
Under matched data, splits, training samples, loss, and compute budget,
RiverFormer outperforms the matched ED-LSTM. On the common five-model
cohort, median KGE increases from $0.65$ to $0.69$ and NSE from $0.54$ to $0.59$.
The station-wise distributions in Figure~\ref{fig:dist-ed-lstm} show a broad shift
toward higher skill rather than an improvement confined to a few gauges.

\begin{figure*}[t]
    \centering
    \includegraphics[width=0.8\textwidth]{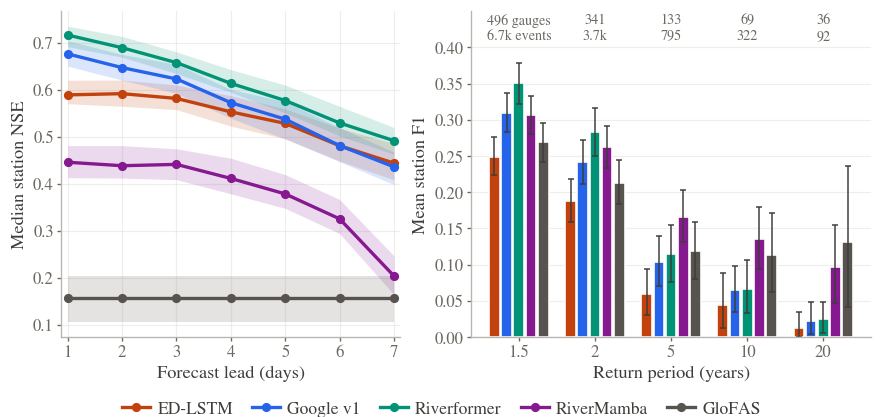}
    \caption{Skill on the paired cohort of 1,033 gauges. Left: median station NSE by
forecast lead. Right: mean station F1 by return period; header rows give the eligible
gauges and observed flood days. Intervals are 95\% percentile bootstrap CIs over gauges, 5,000
replicates.}
    \label{fig:metrics}
\end{figure*}

\textbf{Published operational benchmarks.}
Over the common five-model cohort, RiverFormer achieves the highest
aggregate KGE/NSE ($0.693/0.594$), compared with $0.679/0.555$
for Google v1 and $0.576/0.362$ for RiverMamba
(Appendix Table~\ref{tab:google_v1_window_results}).
RiverFormer has the highest median
NSE point estimate at every lead (Figure~\ref{fig:metrics}, left), although its bootstrap confidence intervals overlap
those of Google v1, indicating comparable skill rather than conclusive
superiority. This is
a conservative comparison because Google v1 was trained on $5{,}680$ gauges,
compared with $3{,}616$ for RiverFormer, and additionally uses snowfall.
RiverFormer has the highest flood F1 point estimate at the 1.5- and 2-year
thresholds (Figure~\ref{fig:metrics}, right). RiverMamba leads at 5 and 10 years and GloFAS at 20 years, but
uncertainty widens as the event count falls from $6{,}718$ to $92$. Moreover,
RiverMamba is not strictly comparable because it consumes GloFAS reanalysis
discharge; its rare-event results are therefore interpreted as a complementary
flood benchmark.

\textbf{Regional performance.}
RiverFormer has the highest NSE in Africa, Europe,
and North America. European NSE and North
American KGE remain close to Google v1. Since North America contributes
738 of the 1,033 gauges, the aggregate results primarily reflect that region
(Appendix Table~\ref{tab:google_v1_window_results}).

 \textbf{Conclusion.}
 Our results show that recurrence is not a prerequisite for operational streamflow forecasting. Using patch-based representations of catchment history and lead-specific conditioning on future meteorology, fully parallel attention can learn rainfall-to-runoff dynamics and match or outperform recurrent approaches. RiverFormer thus offers a foundation architecture for river forecasting that can be extended beyond lumped catchment inputs to models attending directly over meteorological grids and resolving where and when hydrological signals emerge.

\newpage
\bibliography{references}
\bibliographystyle{plainnat}
\small

\appendix

\section{Dataset Details}
\label{app:data}

\subsection{Gauge selection.}
We begin from the $5{,}680$ gauges
released with the operational ED-LSTM of~\citet{nearing2024global} and retain
those we can both force and supervise (Table~\ref{tab:funnel}). Starting from
the baseline's own training population guarantees that no gauge enters our
training set that the baseline did not also see.

\begin{table}[h]
\centering\small
\caption{From the published baseline population to our evaluation cohort.}
\label{tab:funnel}
\begin{tabular}{lrr}
\toprule
step & gauges & $\Delta$ \\
\midrule
published with the operational ED-LSTM & $5{,}680$ & \\
present in the Caravan v1.6 GRDC extension & $3{,}774$ & $-1{,}906$ \\
with $\geq 365$ days of pre-2016 streamflow & $3{,}765$ & $-9$ \\
after flatline and variance quality control & $\mathbf{3{,}616}$ & $-149$ \\
\bottomrule
\end{tabular}
\end{table}

The $1{,}906$ gauges lost at the Caravan step are gauges for which the
baseline publishes hindcasts but which are absent from the Caravan v1.6 GRDC
extension. Without a Caravan entry there is no catchment polygon, and hence no
basin-averaged meteorological forcing, so these gauges cannot be run through
our model.

The final step removes gauges whose observed record fails two standard quality
checks, both evaluated over all non-missing daily values at that gauge. The
flatline ratio, defined as the fraction of consecutive daily values that are
exactly equal, must be below $0.9$, which discards records dominated by
repeated or held-constant readings. The variance of the record must be at least
$10^{-3}$\,($\text{mm}^3$\,d$^{-1}$)$^2$, which discards near-constant series that carry
no hydrological signal. Together these remove $149$ gauges, leaving the
$3{,}616$-gauge training and model-development cohort. Comparisons with
published forecasts use smaller intersections determined by their gauge
coverage, as detailed in Appendix~\ref{app:training}.

\begin{table}[h]
    \centering
    \caption{Basin-averaged daily meteorological input variables and their
    model-ready units, following \citet{taccari2026aifl}.}
    \label{tab:dynamic_variables}
    \begin{tabular}{llll}
        \toprule
        Variable & Description & Unit & Model component \\
        \midrule
        SSR  & Surface net solar radiation   & $\mathrm{W\,m^{-2}}$       & Encoder and decoder \\
        STR  & Surface net thermal radiation & $\mathrm{W\,m^{-2}}$       & Encoder and decoder \\
        SP   & Surface pressure              & $\mathrm{kPa}$              & Encoder and decoder \\
        T2M  & Air temperature at $2$ metres & ${}^{\circ}\mathrm{C}$      & Encoder and decoder \\
        TP   & Total daily precipitation     & $\mathrm{mm\,d^{-1}}$       & Encoder and decoder \\
        CPC  & Total daily precipitation     & $\mathrm{mm\,d^{-1}}$       & Encoder only \\
        IMERG & Total daily precipitation    & $\mathrm{mm\,d^{-1}}$       & Encoder only \\
        \bottomrule
    \end{tabular}
\end{table}
\subsection{Variables.}
Table~\ref{tab:dynamic_variables} summarizes the basin-averaged daily
meteorological inputs. We use the same five ECMWF forcing variables as
\citet{nearing2024global} and \citet{taccari2026aifl}. The model of
\citet{nearing2024global} additionally uses CPC and IMERG precipitation and
snowfall. We also include CPC and IMERG, whereas snowfall is unavailable to
both our model and \citet{taccari2026aifl} and is therefore excluded.

The encoder and decoder use the same five ECMWF meteorological variables.
For each day in the $K=365$-day encoder history, we use the shortest-lead
HRES forecast (lead one) whenever it is available. This provides an
operational estimate of the historical meteorological state without relying
on future information. The HRES archive begins on 2016-01-01; for earlier
training dates, we therefore use ERA5-Land~\citep{era5} as a
fallback for the same five variables. Consequently, ERA5-Land is used only
to extend the training record into the pre-HRES period, whereas the
evaluation periods are driven entirely by HRES. This construction ensures
that evaluation uses only meteorological information that would have been
available operationally at forecast time.

For future forcing, the decoder receives the five HRES variables at leads
$1$--$7$ together with a binary availability indicator, following
\citet{nearing2024global}. CPC and IMERG precipitation are included as
additional historical encoder channels, each accompanied by a daily
availability indicator so that the model can distinguish observations from
fallback values. Their records begin in $1979$ and $2000$, respectively.
Static conditioning uses $84$ Caravan catchment attributes, following
\citet{nearing2024global}. Overall, the resulting input configuration closely
matches that of \citet{nearing2024global}, with snowfall being the main
meteorological variable available to their model but not to ours. Their
published baseline also uses a larger training-gauge population, making the
comparison conservative with respect to both input information and training
data.

\subsection{Data assembly and time convention.}

All datasets are aligned by Caravan gauge identifier and daily date. Streamflow is taken from Caravan v1.6, while ERA5-Land, HRES, CPC, and IMERG are taken from Caravan MultiMet, which provides basin-averaged meteorological variables for each catchment. For each forecast sample, we define a cutoff day $I$: all historical inputs are restricted to values dated on or before $I$, and the model predicts discharge for day $I$ (nowcast) and days $I+1,\ldots,I+7$.

The future meteorological inputs come from the HRES forecast associated with the following day. Under the MultiMet convention, forecast lead $L$ is valid for day $I+L$ and is paired with the discharge observation from that day. Thus, no meteorological value valid after the corresponding prediction day is used.

MultiMet meteorological variables are aggregated in UTC, whereas Caravan streamflow retains the daily convention of its original source dataset. Consequently, matching date labels may not always correspond to exactly the same 24-hour aggregation window. In addition, MultiMet does not provide historical data-arrival timestamps or archived real-time product versions. Our evaluation therefore reproduces the information available by calendar date, but does not attempt to reproduce the exact timing of a real-time operational data-ingestion system.

\begin{figure}[t]
\centering
\resizebox{\linewidth}{!}{%
\begin{tikzpicture}[
  font=\sffamily,
  rowlab/.style={font=\footnotesize\bfseries, anchor=east, align=right},
  tick/.style  ={font=\scriptsize, text=black!65, anchor=south},
  bar/.style   ={rounded corners=1.5pt},
  inbar/.style ={font=\scriptsize, align=center},
  inbarw/.style={font=\scriptsize, text=white, align=center},
  cell/.style  ={font=\tiny, text=white},
  every node/.append style={inner sep=1.5pt}
]
\definecolor{cFallback}{HTML}{DBE6EF}
\definecolor{cHres}{HTML}{2374A5}
\definecolor{cPrecip}{HTML}{E3E9DF}
\definecolor{cFcst}{HTML}{D9822B}
\definecolor{cTarget}{HTML}{7353A4}
\definecolor{cStatic}{HTML}{EDE9F4}

\def\xA{0}\def\xH{2.60}\def\xI{7.20}\def\xE{11.60}
\pgfmathsetmacro{\dw}{(\xE-\xI)/7}

\fill[black!4] (\xI,1.25) rectangle (\xE,5.35);

\draw[-{Stealth[length=2mm]}, gray!70, line width=.45pt] (0,5.66) -- (12,5.66);
\node[tick] at (\xA,5.92) {$I-364$};
\node[tick] at (\xE,5.92) {$I+7$};
\node[font=\footnotesize\bfseries, anchor=south] at (\xI,5.92) {$I$};
\draw[black!85, line width=.8pt] (\xI,1.20) -- (\xI,5.66);

\draw[decorate, decoration={brace, amplitude=3pt, mirror}, gray!70] (\xA,5.55) -- (\xI-.03,5.55);
\draw[decorate, decoration={brace, amplitude=3pt, mirror}, gray!70] (\xI+.03,5.55) -- (\xE,5.55);
\node[font=\footnotesize, anchor=north] at ({(\xA+\xI)/2},5.45) {encoder: 365 days};
\node[font=\footnotesize, anchor=north] at ({(\xI+\xE)/2},5.45) {decoder: HRES run $T=I{+}1$};

\node[rowlab] at (-0.15,4.35) {ECMWF $\times5$};
\path[bar, fill=cFallback] (\xA,4.08) rectangle (\xH,4.63);
\path[bar, fill=cHres]     (\xH,4.08) rectangle (\xI,4.63);
\node[inbar]  at ({(\xA+\xH)/2},4.35) {ERA5-Land};
\node[inbarw] at ({(\xH+\xI)/2},4.35) {HRES lead 1};
\path[bar, fill=cFcst] (\xI,4.08) rectangle (\xE,4.63);
\foreach \k in {1,...,7}{%
  \pgfmathsetmacro{\bx}{\xI+\k*\dw}
  \ifnum\k<7\draw[white, line width=.6pt] (\bx,4.08) -- (\bx,4.63);\fi
  \node[cell] at ({\bx-\dw/2},4.35) {$L{=}\k$};}

\node[rowlab] at (-0.15,3.25) {CPC $+$ IMERG};
\path[bar, preaction={fill=cPrecip}, pattern=north east lines,
      pattern color=green!30!black!35] (\xA,2.98) rectangle (\xI,3.53);
\node[inbar] at ({(\xA+\xI)/2},3.25) {observed precipitation};

\node[rowlab] at (-0.15,2.28) {Static};
\path[bar, fill=cStatic] (\xA,2.06) rectangle (\xE,2.50);
\node[inbar] at ({(\xA+\xE)/2},2.28) {84 catchment attributes};

\node[rowlab] at (-0.15,1.60) {Targets};
\path[bar, fill=cTarget] ({\xI-\dw},1.33) rectangle (\xI,1.88);
\path[bar, fill=cTarget] (\xI,1.33) rectangle (\xE,1.88);
\draw[white, line width=.9pt] (\xI,1.33) -- (\xI,1.88);
\node[cell] at ({\xI-\dw/2},1.60) {$\tau{=}0$};
\foreach \k in {1,...,7}{%
  \pgfmathsetmacro{\bx}{\xI+\k*\dw}
  \ifnum\k<7\draw[white, line width=.6pt] (\bx,1.33) -- (\bx,1.88);\fi
  \node[cell] at ({\bx-\dw/2},1.60) {$\tau{=}\k$};}
\node[font=\scriptsize, text=black!65, anchor=north] at ({\xI-\dw/2},1.26) {nowcast};
\node[font=\scriptsize, text=black!65, anchor=north] at ({(\xI+\xE)/2},1.26) {forecasts};
\end{tikzpicture}%
}
\caption{Calendar-causal assembly of one forecast. }
\label{fig:data-merging-timeline}
\end{figure}
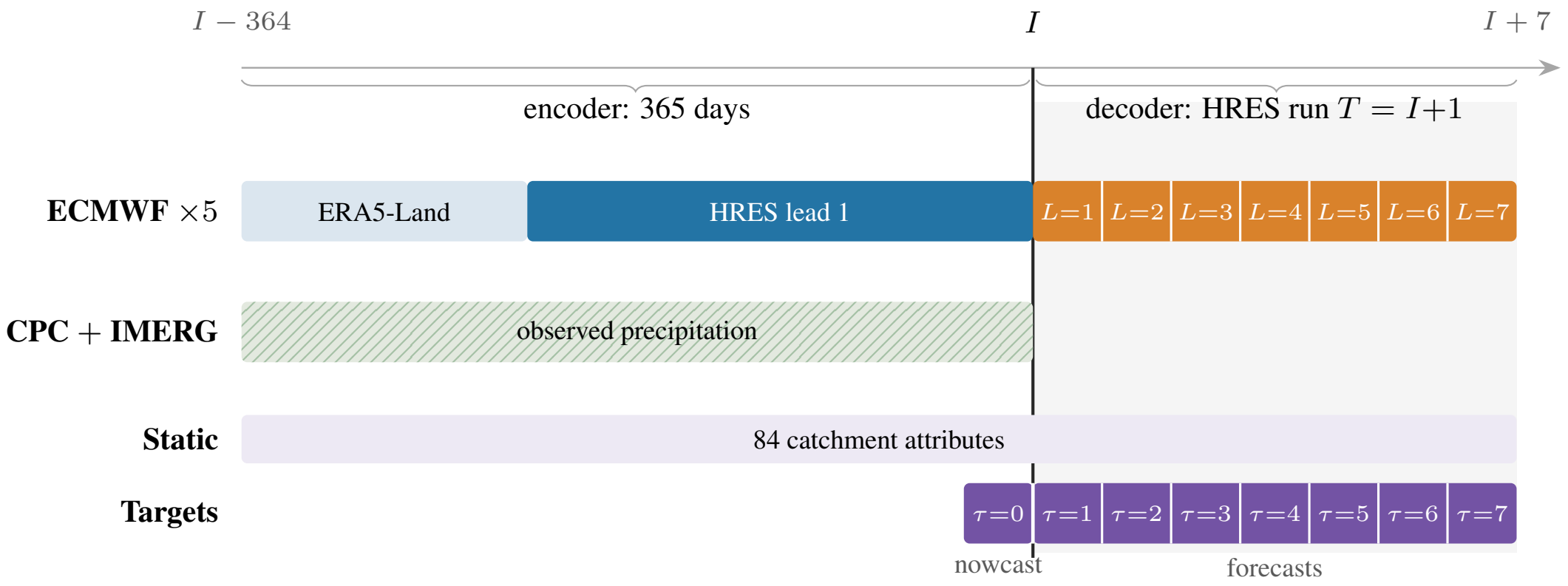

\subsection{Forecast horizons.}
For cutoff day $I$, the encoder receives a $365$-step meteorological history
containing no values valid after $I$. The decoder receives a single HRES forecast
with nominal base date $T=I+1$ and leads valid on $I+1,\ldots,I+7$. The model
emits $\tau=0,\ldots,7$, where $\tau=0$ is a nowcast for day $I$ and
$\tau=1,\ldots,7$ are forecasts for days $I+\tau$. For comparison with the
released Google archive, we map archive timestamp $T$ to cutoff date $I=T-1$;
archive lead $\tau$ is consequently scored against day
$T-1+\tau=I+\tau$. This alignment is applied consistently so that all
comparisons are made for identical target days and lead indices.

\subsection{Normalisation.}
All dynamic input channels are standardised to zero mean and unit variance using statistics computed from the training period. For each channel, a single mean and standard deviation are estimated across all training gauges and timesteps while ignoring missing values. Standard deviations smaller than $10^{-6}$ are replaced by one. For the five ECMWF weather variables, normalisation is fitted to the actual combined input seen by the model: HRES where available and ERA5-Land otherwise. Since both products are provided in the same physical units, they are normalised jointly rather than separately. This avoids introducing an artificial shift when the input source changes from ERA5-Land to HRES. The same procedure is used for CPC and IMERG precipitation, with ERA5-Land precipitation serving as the fallback. Static catchment attributes are median-imputed per feature and then standardised across gauges. Streamflow targets are globally standardised using the  discharge values from the training period.

\section{Experimental Details}
\label{app:training}

\subsection{Splits and benchmarking}

Our evaluation is gauged out-of-time. Training is restricted to $1980$-$01$-$01$--$2019$-$10$-$24$, followed by a one-year validation period from $2019$-$10$-$25$ to $2020$-$10$-$24$ used for model and checkpoint selection; all observations from $2020$-$10$-$25$ onward are reserved for testing. We use a single chronological split, similar to \citet{taccari2026aifl}, such that every test observation occurs strictly after the data available for model training and selection. In contrast, \citet{nearing2024global} train for a fixed set of $50{,}000$ minibatches rather than selecting the final checkpoint using a held-out temporal validation period.

A fully fair comparison with the published forecasts of \citet{nearing2024global} is not possible. Their model was trained on a larger gauge set and additionally uses snowfall, which is unavailable in the public dataset. Their reported temporal benchmark also relies on temporal cross-validation, whereas we use a single future holdout. Temporal cross-validation provides predictions over a longer evaluation period, but for an earlier test year the corresponding training fold may contain observations from later years. This is a valid cross-validation protocol, but it can expose the model to later hydrological regimes, for example changes caused by regulation, land use, or climate. Reproducing the full protocol would also be computationally expensive: for eight evaluation years, a three-member ensemble would require $8\times3=24$ separate training runs per architecture.

We therefore report two complementary comparisons. First, we compare
RiverFormer with the published temporal-cross-validation forecasts of
\citet{nearing2024global} over their overlapping period from October 2020 to
October 2021. Of the $3{,}616$ gauges used to develop RiverFormer, $1{,}663$ also
have Google v1 forecasts in this window. Requiring predictions from RiverMamba
and GloFAS further reduces the common five-model intersection to $1{,}033$
gauges; this is the cohort used in Table~\ref{tab:google_v1_window_results} so
that every entry is evaluated on identical stations and dates. These counts
are evaluation intersections, not training-set sizes.

Second, for a controlled architecture comparison, we train the
ED-LSTM using exactly the same gauges, inputs, temporal
split, and optimization constraints as RiverFormer and evaluate both models on the five-model cohort in the Google v1 window (Table~\ref{tab:google_v1_window_results}) and over the extended test period through May 2023 (Table~\ref{tab:extended_window_results}). Since the
operational ED-LSTM is reported as a three-member ensemble~\citep{nearing2024global},
we follow the same protocol for both architectures: three independently
initialized models are trained and their point predictions are averaged. For
RiverFormer, this is equivalent to the mean prediction of an equally weighted
pooling of the three predictive mixtures. The published comparison anchors our
results to an operational baseline, while the matched ED-LSTM isolates the
effect of replacing recurrence with RiverFormer.

AIFL~\citep{taccari2026aifl} is discussed as related operational work but is
not included as a quantitative baseline because its trained model and
station-level predictions had not been released at the time of this study.

\subsection{Flood detection}

We evaluate extreme-event skill following RiverMamba~\citep{RiverMamba}.
For each gauge we fit a Gumbel distribution to the calendar-year maxima of the
observed record by L-moments and take return levels at 1.5, 2, 5, 10 and 20 years
as flood thresholds, fitting only on the training and validation span
(1980-01-01 to 2020-10-24) so that no threshold has seen a scored day.
Detection is a per-day classification over the held-out test year: a day is
positive when the observation exceeds the threshold, and the model scores a true
positive when its prediction exceeds the \emph{same} threshold. We report
F1 per gauge and lead time, averaged over gauges. \citet{nearing2024global} instead fit a separate
distribution to each modelled series and match rising-edge events within a
tolerance window. Under a single temporal split it would calibrate each threshold on years the model was trained on and measure the test year against an inflated bar, whereas 
observation-only thresholds place every model on the same bar, so that models have to predict not only high flow but also the severity of the peak. Per-day classification is
also better conditioned here, as a one-year window leaves too few event
crossings per gauge for stable event-matched scores.

\subsection{Compute and Optimization}

RiverFormer and the ED-LSTM baseline use the same data, splits, training samples, CMAL output head, loss, and target noise. Both are trained for $300$k steps with batch size $2048$, a $9$k-step warmup followed by a stable phase and a $100$k-step cosine decay, gradient clipping at $1.0$, and multiplicative target noise with standard deviation $0.005$. Checkpoints are selected using continent-median NSE on the validation period (2019-10-25--2020-10-24), which is disjoint from the reported test period. RiverFormer has approximately $2.3$ million trainable parameters. Training
each seed for $300$k steps takes approximately three hours on a single NVIDIA
GeForce RTX 5090 GPU.

\begin{table}[t]
\centering
\small
\setlength{\tabcolsep}{4pt}
\caption{Final RiverFormer configuration. Reported results use a three-member ensemble whose point predictions are averaged.}
\label{tab:riverformer-config}
\begin{tabular}{@{}l p{0.63\linewidth}@{}}
\toprule
Component & Setting \\
\midrule
History tokenization & 4-day patches, 91 patches over a 365-day lookback \\
History encoder      & 3 layers, 4 heads, width 64, FFN 256, GELU, pre-norm \\
Static encoder       & MLP $84\!\rightarrow\!100\!\rightarrow\!100\!\rightarrow\!8$, prepended to the history tokens as a context token \\
Forecast decoder     & 2 layers, 4 heads, width 256, FFN 1024, GELU, pre-norm, cross-attention over the encoder memory \\
Decoder queries      & Projected weather forcings plus a learned nowcast query, with additive lead embeddings (leads 0--7) \\
Dropout              & 0.1 \\
\midrule
Optimizer            & Muon (Newton--Schulz, 5 steps) for 2-D weights, fused AdamW otherwise \\
Muon LR              & $5\times10^{-4}$, momentum 0.95, Nesterov, weight decay 0.01 \\
AdamW LR             & $2\times10^{-4}$, $\beta=(0.9,0.95)$, weight decay 0 \\
Schedule             & Warmup--stable--cosine: 9k warmup, 191k stable, 100k decay to $0.1\times$ \\
Steps / batch        & 300{,}000 / 2048 \\
Grad clip            & 1.0 \\
Precision            & bf16 \\
\bottomrule
\end{tabular}
\end{table}

RiverFormer uses Muon~\citep{jordan2024muon} (learning rate $5\times10^{-4}$) for matrix parameters and AdamW~\citep{loshchilov2019adamw} (learning rate $2\times10^{-4}$) for the remaining parameters, and is trained in BF16 with compilation. The reported ED-LSTM (M1, below) also uses Muon with auxiliary AdamW but is trained in FP32.

\paragraph{Ensuring parity with the LSTM baseline.}
The original ED-LSTM recipe uses batch size 256 and a fixed training budget.
Directly comparing this recipe with our training setup would confound
architecture and optimization. We therefore evaluate five optimization
settings, keeping the ED-LSTM architecture fixed within each implementation:
\begin{itemize}
  \item A0: our ported ED-LSTM with AdamW, batch size 256 and constant
        learning rate $10^{-3}$;
  \item A1: the same ported architecture and batch size with our
        warmup--stable--cosine learning-rate schedule;
  \item A2: the same ported architecture with batch size 2048, our
        learning-rate schedule, and learning rate $5\times10^{-4}$;
  \item M0: the same ported ED-LSTM with Muon and auxiliary AdamW,
        batch size 256, our learning-rate schedule, and Muon learning
        rate $5\times10^{-4}$;
  \item M1: the same ported architecture and optimizer settings
        with batch size 2048, matching RiverFormer in batch size and
        nominal training-sample budget.
\end{itemize}
All five settings use a 300k-step budget and seeds 2024--2026.
A0--A2 use zero weight decay, so their AdamW updates are equivalent to
Adam updates. All other ED-LSTM settings, including the architecture,
CMAL head, initialization, dropout, target noise, and input data, remain
unchanged within A0--A2 and within M0--M1. We report M1 because it matches RiverFormer in batch size and training
budget. M0 and M1 perform similarly, with M0 obtaining slightly higher
aggregate scores (Table~\ref{tab:edlstm-batch-ablation}). These results do
not establish a performance benefit from the larger batch.

\begin{table}[t]
\centering
\caption{ED-LSTM optimization comparison over three seeds (2024--2026).
Checkpoints are selected by NSE on the validation period
(2019-10-25--2020-10-24); the test window is 2020-10-25--2021-10-24.}
\label{tab:edlstm-batch-ablation}
\begin{tabular}{lrrr}
\toprule
Setting & Batch & Test $\mathrm{KGE}$ & Test NSE \\
\midrule
A0 (AdamW, constant LR) & 256 & 0.572 & 0.446 \\
A1 (AdamW, scheduled) & 256 & 0.563 & 0.440 \\
A2 (AdamW, scheduled) & 2048 & 0.608 & 0.490 \\
M0 (Muon + AdamW) & 256 & 0.647 & 0.538 \\
M1 (Muon + AdamW, reported) & 2048 & 0.645 & 0.537 \\
\bottomrule
\end{tabular}
\end{table}

Several implementation differences remain between RiverFormer and the
reported M1 baseline. Both models use Muon for
selected matrix parameters and AdamW for the remaining parameters.
For the ED-LSTM, Muon is applied to the recurrent weight matrices and
selected dense projections, while the output-head parameters use AdamW.
The two architectures differ in numerical precision. Although BF16 execution
is supported by modern PyTorch/CUDA backends, its behavior for recurrent
models depends on the underlying cuDNN implementation and does not
necessarily provide the same stability or efficiency as for Transformer
workloads. We train the baseline in FP32, whereas RiverFormer is trained
in BF16 with compilation.

\section{Extended Results}

\subsection{Continent View and Extended Test Window}
As shown in Tables \ref{tab:google_v1_window_results} and  \ref{tab:extended_window_results}, the extended window preserves the main aggregate ranking:
RiverFormer outperforms ED-LSTM, RiverMamba, and GloFAS on both KGE and NSE.
Its advantage over the matched ED-LSTM also remains stable: the KGE/NSE gaps change from $0.048/0.057$ in 2020--2021 to $0.045/0.059$ over 2020--2023.
Over the extended window, RiverFormer leads both metrics in Africa,
Europe, North America, and Oceania. The changing South American ranking should be interpreted
cautiously because the cohort contains only eight or nine gauges. Although all
four models obtain higher aggregate scores over the extended window, this
likely reflects differences in the evaluated years rather than model
improvement. Google v1 is omitted because forecasts are unavailable beyond
2021.

\begin{table}[h]
    \centering
    \caption{Median station KGE and NSE by continent over the common five-model
    intersection during the Google v1 evaluation window.}
    \label{tab:google_v1_window_results}
    \resizebox{\linewidth}{!}{%
        \begin{tabular}{lr|rrrrr|rrrrr}
            \toprule
            &
            & \multicolumn{5}{c|}{KGE}
            & \multicolumn{5}{c}{NSE} \\
            \cmidrule(lr){3-7}
            \cmidrule(l){8-12}
            Continent
            & $n$
            & ED-LSTM & Google v1 & RiverFormer & RiverMamba & GloFAS
            & ED-LSTM & Google v1 & RiverFormer & RiverMamba & GloFAS \\
            \midrule

            Africa
            & 102
            & 0.356 & 0.396 & \textbf{0.459} & 0.277 & 0.161
            & 0.280 & 0.258 & \textbf{0.355} & -0.555 & -0.110 \\

            Europe
            & 124
            & 0.787 & 0.751 & \textbf{0.799} & 0.696 & 0.474
            & 0.715 & 0.712 & \textbf{0.743} & 0.534 & 0.125 \\

            North America
            & 738
            & 0.660 & 0.711 & \textbf{0.713} & 0.597 & 0.465
            & 0.544 & 0.591 & \textbf{0.623} & 0.397 & 0.193 \\

            Oceania
            & 61
            & 0.433 & 0.422 & \textbf{0.515} & 0.504 & 0.357
            & \textbf{0.388} & 0.202 & 0.350 & 0.250 & 0.209 \\

            South America
            & 8
            & 0.489 & 0.543 & \textbf{0.622} & 0.606 & 0.544
            & 0.188 & 0.290 & 0.493 & \textbf{0.612} & -0.241 \\

            \midrule

            ALL
            & 1{,}033
            & 0.645 & 0.679 & \textbf{0.693} & 0.576 & 0.435
            & 0.537 & 0.555 & \textbf{0.594} & 0.362 & 0.156 \\

            \bottomrule
        \end{tabular}%
    }
\end{table}

\begin{table}[h]
    \centering
    \caption{Median station KGE and NSE by continent over the extended evaluation window
    (October 2020--May 2023). The best-performing value for each metric in each row is shown in bold.}
    \label{tab:extended_window_results}
    \resizebox{\linewidth}{!}{%
        \begin{tabular}{lr|rrrr|rrrr}
            \toprule
            &
            & \multicolumn{4}{c|}{KGE}
            & \multicolumn{4}{c}{NSE} \\
            \cmidrule(lr){3-6}
            \cmidrule(l){7-10}
            Continent
            & $n$
            & ED-LSTM & RiverFormer & RiverMamba & GloFAS
            & ED-LSTM & RiverFormer & RiverMamba & GloFAS \\
            \midrule

            Africa
            & 104
            & 0.354 & \textbf{0.447} & 0.432 & 0.371
            & 0.211 & \textbf{0.319} & -0.055 & 0.045 \\

            Europe
            & 128
            & 0.808 & \textbf{0.813} & 0.696 & 0.476
            & 0.751 & \textbf{0.765} & 0.559 & 0.207 \\

            North America
            & 740
            & 0.695 & \textbf{0.742} & 0.646 & 0.532
            & 0.608 & \textbf{0.676} & 0.483 & 0.317 \\

            Oceania
            & 61
            & 0.512 & \textbf{0.565} & 0.515 & 0.391
            & 0.459 & \textbf{0.493} & 0.331 & 0.236 \\

            South America
            & 9
            & 0.331 & 0.582 & 0.595 & \textbf{0.615}
            & 0.142 & 0.469 & \textbf{0.575} & -0.217 \\

            \midrule

            ALL
            & 1042
            & 0.680 & \textbf{0.725} & 0.625 & 0.488
            & 0.584 & \textbf{0.643} & 0.438 & 0.258 \\

            \bottomrule
        \end{tabular}%
    }
\end{table}

\subsection{Ablations}
\textbf{HRES-era oversampling.}
Operational HRES forecasts are available only from 2016 onward and therefore
account for less than ten percent of the 1980--2019 training period. Under
uniform sampling, the decoder consequently encounters ERA5-Land fallback
forcings much more frequently than genuine HRES forecasts. Whereas
\citet{taccari2026aifl} address this imbalance through a separate HRES
fine-tuning stage, we evaluated a single-stage alternative based on stratified
sampling. A fraction $f_{\mathrm{HRES}}$ of each batch is drawn from
issue-date--gauge pairs with complete HRES forecasts at every lead, while the
remainder is sampled from the full training set. Contrary to our expectation, increasing $f_{\mathrm{HRES}}$ consistently
reduces validation performance (Table~\ref{tab:sampling_results}). We
hypothesize that repeatedly sampling the relatively short HRES period reduces
the diversity of hydrological conditions encountered during training, and
that this loss outweighs the benefit of greater exposure to the operational
forcing distribution. We therefore use $f_{\mathrm{HRES}}=0$ for all main
results. This setting corresponds to uniform sampling rather than the complete
exclusion of HRES-era examples, which remain present at their natural
frequency. 

\begin{table}[h]
    \centering
    \caption{KGE and NSE HRES sampling performance by continent, scored on the validation split ( 2019-10-25--2020-10-24) at a fixed 100k-step budget. The best-performing value for each metric in each row is shown in bold.}
    \label{tab:sampling_results}
    \begin{tabular}{lrrrr|rrr}
        \toprule
        & & \multicolumn{3}{c|}{KGE} & \multicolumn{3}{c}{NSE} \\
        \cmidrule(lr){3-5} \cmidrule(l){6-8}
        Continent & $n$ & 0.0 & 0.5 & 1.0 & 0.0 & 0.5 & 1.0 \\
        \midrule
            Africa
                & 196
                & 0.3163 & 0.3309 & \textbf{0.3603}
                & 0.1899 & \textbf{0.2154} & 0.2073 \\
            Europe
                & 334
                & \textbf{0.7772} & 0.7582 & 0.7347
                & \textbf{0.7258} & 0.7090 & 0.6524 \\
            North America
                & 1329
                & \textbf{0.7559} & 0.7442 & 0.7200
                & \textbf{0.6992} & 0.6872 & 0.6552 \\
            Oceania
                & 156
                & \textbf{0.4603} & 0.4549 & 0.4541
                & \textbf{0.3344} & 0.3067 & 0.3308 \\
            South America
                & 156
                & \textbf{0.6760} & 0.6586 & 0.6483
                & \textbf{0.5968} & 0.5702 & 0.5582 \\
        \midrule
            ALL
                & 2171
                & \textbf{0.7167} & 0.7112 & 0.6878
                & \textbf{0.6503} & 0.6338 & 0.6046 \\
        \bottomrule
    \end{tabular}
\end{table}

\textbf{Patch-length sensitivity.}
We performed a single-seed screening experiment over patch lengths
$p\in\{1,4,16,32\}$ using a truncated 100k-step training budget
(Table~\ref{tab:patch_results}). All patched configurations outperform the
unpatched daily-token configuration ($p=1$) in aggregate, supporting the use of
temporal patching. Under this reduced training regime, $p=16$ obtains the
highest aggregate validation KGE and NSE. However, the ranking varies across
continents, and this experiment neither uses the full 300k-step training
schedule nor estimates variability across random seeds. The main experiments
use $p=4$; in the corresponding full-budget runs, its three-seed ensemble
performed slightly better than the $p=16$ ensemble. Repeating
the complete four-value sweep with three seeds at 300k steps would require 12
full training runs and
is computationally expensive. We therefore interpret the low-budget sweep
as evidence for the benefit of patching.

\begin{table}[h]
    \centering
    \caption{Single-seed KGE and NSE performance by continent and patch size, scored on the validation split (2019-10-25--2020-10-24) at a fixed 100k-step budget with $f_{\mathrm{HRES}}=0$ (uniform sampling).
    The best-performing value for each metric in each row is shown in bold.}
    \label{tab:patch_results}
    \resizebox{\linewidth}{!}{%
        \begin{tabular}{lrrrrr|rrrr}
            \toprule
            & & \multicolumn{4}{c|}{KGE} & \multicolumn{4}{c}{NSE} \\
            \cmidrule(lr){3-6} \cmidrule(l){7-10}
            Continent & $n$
                & $p=1$ & $p=4$ & $p=16$ & $p=32$
                & $p=1$ & $p=4$ & $p=16$ & $p=32$ \\
            \midrule
            Africa
                & 196
                & 0.2204 & \textbf{0.3244} & 0.3163 & 0.2954
                & 0.0045 & 0.1256 & \textbf{0.1899} & 0.1492 \\
            Europe
                & 334
                & 0.7423 & 0.7521 & \textbf{0.7772} & 0.7555
                & 0.6610 & 0.7038 & \textbf{0.7258} & 0.6901 \\
            North America
                & 1329
                & 0.7246 & 0.7448 & \textbf{0.7559} & 0.7476
                & 0.6331 & 0.6921 & \textbf{0.6992} & 0.6979 \\
            Oceania
                & 156
                & 0.4250 & 0.4452 & 0.4603 & \textbf{0.4617}
                & 0.3544 & \textbf{0.3860} & 0.3344 & 0.3479 \\
            South America
                & 156
                & \textbf{0.6853} & 0.6601 & 0.6760 & 0.6735
                & 0.5704 & 0.5702 & \textbf{0.5968} & 0.5358 \\
            \midrule
            ALL
                & 2171
                & 0.6889 & 0.7102 & \textbf{0.7167} & 0.7145
                & 0.5866 & 0.6358 & \textbf{0.6503} & 0.6367 \\
            \bottomrule
        \end{tabular}%
    }
     \label{table:patch}
\end{table}

\subsection{Prediction Distributions}

\begin{figure*}[h]
    \centering
    \includegraphics[width=0.8\textwidth]{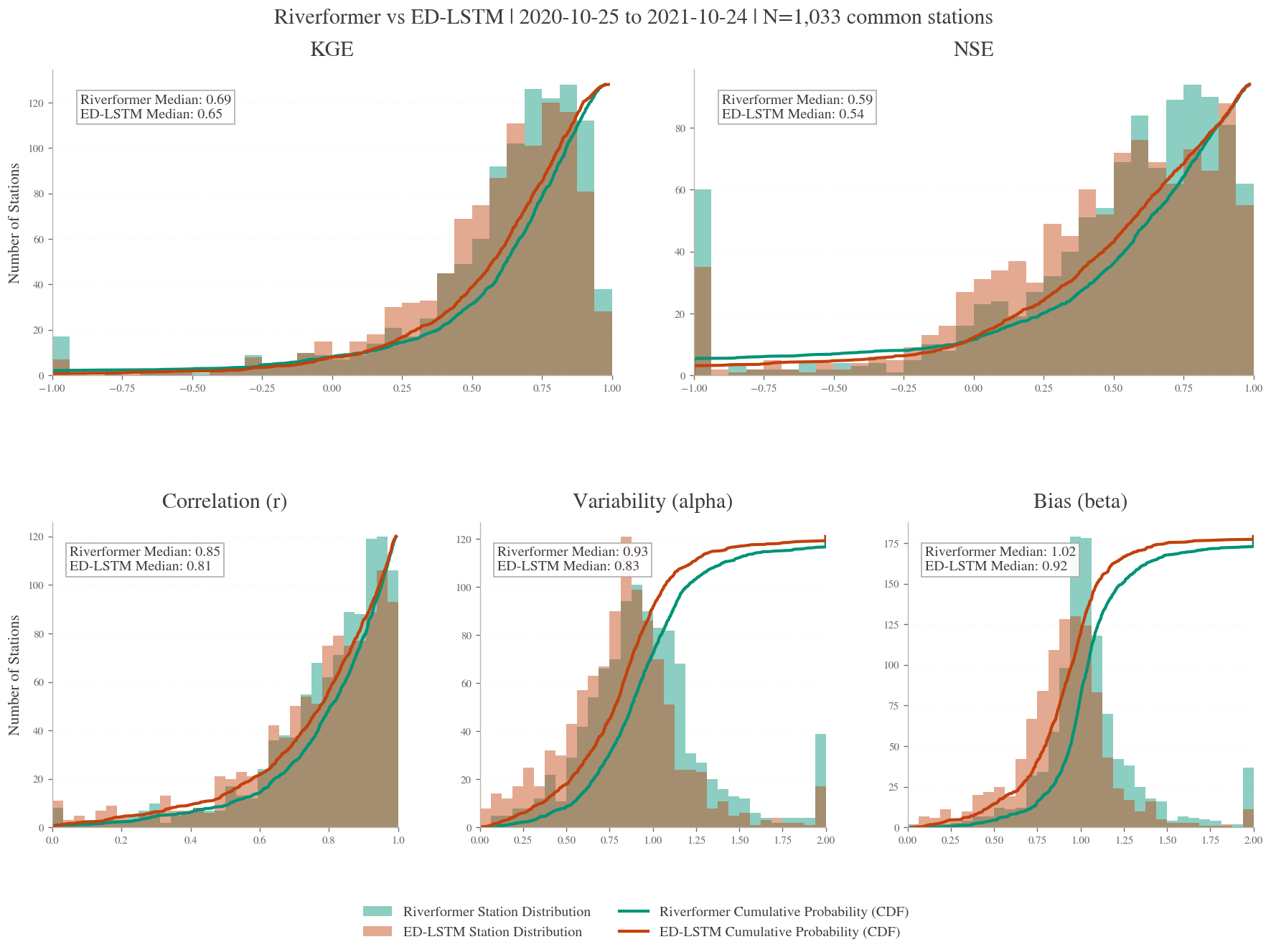}
    \caption{Station-level distributions and empirical CDFs of KGE, NSE, and
their diagnostic components for RiverFormer and the matched ED-LSTM.}
    \label{fig:dist-ed-lstm}
\end{figure*}

\begin{figure*}[h]
    \centering
    \includegraphics[width=0.8\textwidth]{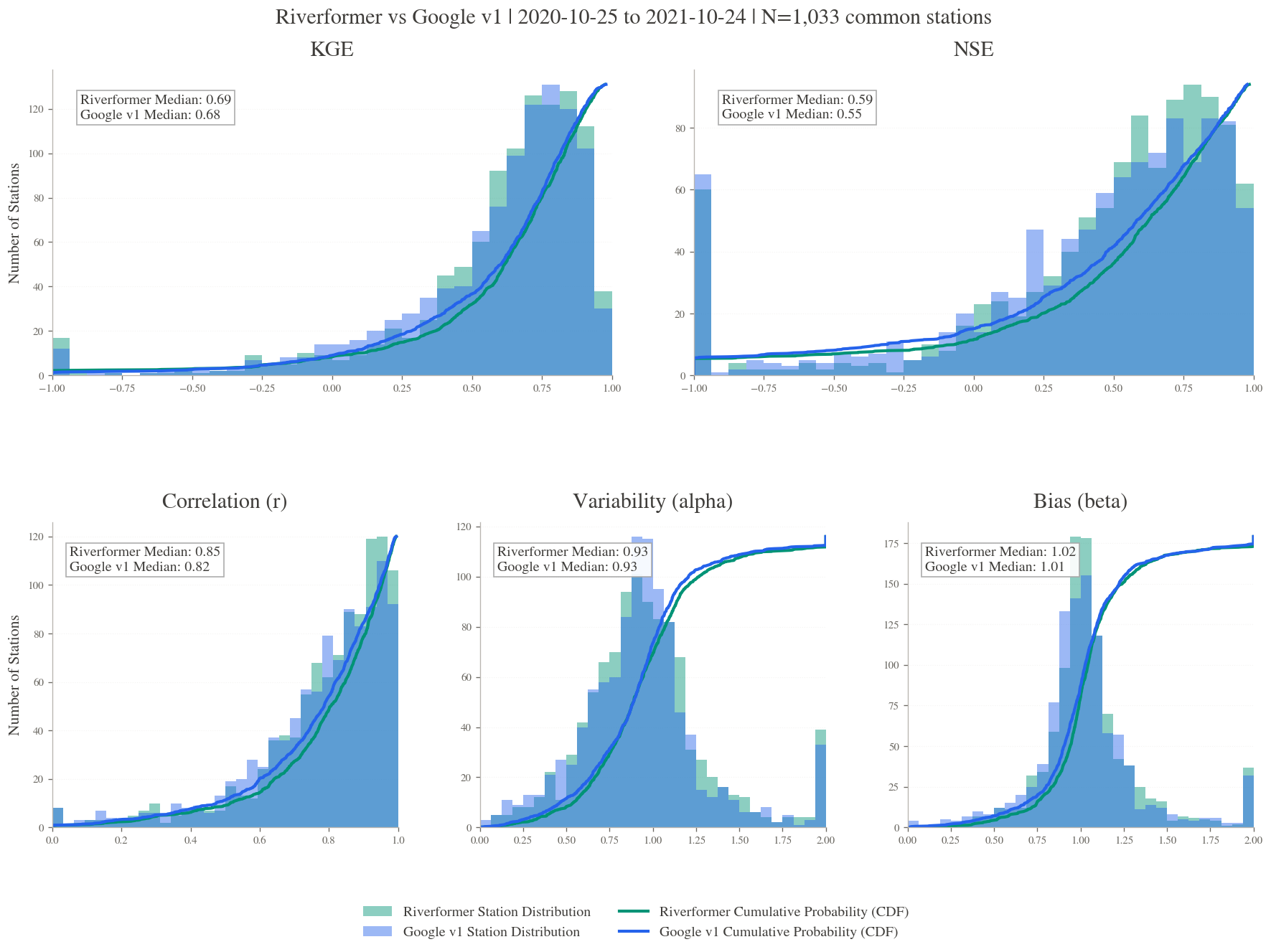}
    \caption{Station-level distributions and empirical CDFs of KGE, NSE, and
their diagnostic components for RiverFormer and published Google v1 forecasts.}
    \label{fig:dist-google}
\end{figure*}

Following the station-wise diagnostic style of
\citet{taccari2026aifl}, Figures~\ref{fig:dist-ed-lstm}
and~\ref{fig:dist-google} show histograms and empirical cumulative distribution
functions (CDFs) of KGE, NSE, and three hydrograph diagnostics across the common
$1{,}033$ gauges. For each gauge, these diagnostics are
\[
r=\operatorname{corr}(x_{\mathrm{sim}},x_{\mathrm{obs}}),
\qquad
\alpha=\frac{\sigma_{\mathrm{sim}}}{\sigma_{\mathrm{obs}}},
\qquad
\beta=\frac{\mu_{\mathrm{sim}}}{\mu_{\mathrm{obs}}},
\]
following the decomposition of \citet{gupta2009kge}. Here $r$ is the Pearson correlation and measures temporal agreement between
predicted and observed discharge, $\alpha$ is the standard-deviation ratio, and
$\beta$ is the mean-flow bias ratio. Note that the reported KGE uses the modified
formulation of \citet{KLING2012264} (Appendix~\ref{app:metrics}), whose variability
term is the coefficient-of-variation ratio $\gamma=\alpha/\beta$ rather than $\alpha$;
we show $\alpha$ because it separates variability errors from mean-bias errors,
which $\gamma$ mixes. $r$ and $\beta$ enter both formulations identically.
For KGE, NSE, and $r$, a distribution shifted toward one indicates better
performance. For $\alpha$ and $\beta$, concentration around one is preferable
to a simple rightward shift.

Figure~\ref{fig:dist-ed-lstm} shows that RiverFormer shifts most
of the KGE and NSE distributions toward higher skill, increasing
their medians from $0.65/0.54$ to $0.69/0.59$ compared to ED-LSTM.
The KGE decomposition shows that this improvement is not driven
only by correlation: median $r$ increases from $0.81$ to $0.85$,
while $\alpha$ moves from $0.83$ to $0.93$ and $\beta$ from
$0.92$ to $1.02$.
The ED-LSTM therefore tends to produce hydrographs with less
variability and underestimated mean flow, whereas RiverFormer is substantially
closer to the ideal variability and bias ratios. The overlapping histograms
and remaining low-skill tail nevertheless show that the improvement is broad
rather than uniform across every gauge.

The comparison with Google v1 is considerably closer, as shown
in Figure~\ref{fig:dist-google}. Their KGE distributions largely
overlap, with medians of $0.69$ for RiverFormer and $0.68$ for
Google v1, while both have identical median $\alpha=0.93$ and
similar median $\beta$ values ($1.02$ and $1.01$, respectively).
RiverFormer has a modestly higher correlation ($0.85$ versus
$0.82$) and NSE ($0.59$ versus $0.55$).
With identical median $\alpha$ and similar median $\beta$,
the modest difference between the models is instead reflected
in RiverFormer's higher median $r$ and NSE, indicating better
temporal agreement with observed discharge and lower squared
prediction error relative to the observed-flow variance.
The strong distributional overlap supports interpreting their
overall performance as comparable rather than claiming
uniform superiority.

\section{Metrics}
\label{app:metrics}

Let $x^t_{\mathrm{obs}}$ be the observed discharge series with mean $\mu_{\mathrm{obs}}$ and standard deviation $\sigma_{\mathrm{obs}}$, and let $x^t_{\mathrm{sim}}$ be the simulated or predicted series with mean $\mu_{\mathrm{sim}}$ and standard deviation $\sigma_{\mathrm{sim}}$.

The Nash--Sutcliffe Efficiency ~\citep{nash1970nse} is
\begin{equation}
    \mathrm{NSE}
    =
    1 -
    \frac{\sum_{t=1}^{n}\left(x^t_{\mathrm{obs}} - x^t_{\mathrm{sim}}\right)^2}
    {\sum_{t=1}^{n}\left(x^t_{\mathrm{obs}} - \mu_{\mathrm{obs}}\right)^2}.
\end{equation}  
NSE compares model squared error against the variance of the observed hydrograph; values above zero improve on predicting the observed mean, and high-flow errors have strong influence.

The Kling--Gupta Efficiency (We used the modified version \cite{KLING2012264}) is
\begin{equation}
    \mathrm{KGE}
    =
    1 -
    \sqrt{
    (1-r)^2
    +
    \left(1-
    \frac{\sigma_{\mathrm{sim}}/\mu_{\mathrm{sim}}}{\sigma_{\mathrm{obs}}/\mu_{\mathrm{obs}}}\right)^2
    +
    \left(1-\frac{\mu_{\mathrm{sim}}}{\mu_{\mathrm{obs}}}\right)^2
    },
\end{equation}  
where $r$ is the Pearson correlation coefficient between the two series. KGE jointly assesses correlation, variability bias, and mean bias, making it complementary to NSE. NSE and KGE are bounded from above by 1, with larger values indicating better forecasts.

\newpage

\end{document}